\documentclass[review]{elsarticle}

\usepackage{hyperref}
\usepackage{graphicx}
\usepackage{amsmath}
\usepackage{amssymb}
\usepackage{multirow}

\begin{document}

\begin{frontmatter}

\title{Class-Aware Domain Adaptation for Improving Adversarial Robustness}



\author[mymainaddress]{Xianxu Hou}
\author[mymainaddress]{Jingxin Liu}
\author[mymainaddress]{Bolei Xu}
\author[sixth]{Xiaolong Wang}
\author[mymainaddress]{Bozhi Liu}
\author[mymainaddress,fifth]{Guoping Qiu}

\address[mymainaddress]{College of Electronic and Information Engineering, Shenzhen University, Shenzhen, China}
\address[sixth]{IBM, San Jose, CA, USA}
\address[fifth]{School of Computer Science, University of Nottingham, Nottingham, United Kingdom}

\begin{abstract}
Recent works have demonstrated convolutional neural networks are vulnerable to adversarial examples, \emph{i.e.}, inputs to machine learning models that an attacker has intentionally designed to cause the models to make a mistake. To improve the adversarial robustness of neural networks, adversarial training has been proposed to train networks by injecting adversarial examples into the training data. However, adversarial training could overfit to a specific type of adversarial attack and also lead to standard accuracy drop on clean images. To this end, we propose a novel Class-Aware Domain Adaptation (CADA) method for adversarial defense without directly applying adversarial training. Specifically, we propose to learn domain-invariant features for adversarial examples and clean images via a domain discriminator. Furthermore, we introduce a class-aware component into the discriminator to increase the discriminative power of the network for adversarial examples. We evaluate our newly proposed approach using multiple benchmark datasets. The results demonstrate that our method can significantly improve the state-of-the-art of adversarial robustness for various attacks and maintain high performances on clean images.
\end{abstract}

\begin{keyword}
Domain adaptation\sep Adversarial Robustness\sep 
\end{keyword}

\end{frontmatter}

\section{Introduction}
Recent works \cite{42503,43405} have shown that deep neural networks are vulnerable to adversarial examples, which are maliciously designed inputs to attack target models by adding small perturbations to clean images. Although the adversarial perturbations are often imperceptible to humans, these adversarial attacks are highly effective against state-of-the-art deep neural networks (DNNs) \cite{krizhevsky2012imagenet,simonyan2015very}. Moreover, adversarial examples can transfer across different models and maintain their effectiveness. The success of adversarial attacks have lead to potential threat for security sensitive deep learning systems. Thus, how to effectively improve adversarial robustness of deep learning models is crucial for real-world applications such as autonomous driving \cite{sitawarin2018darts} as well as identity authentication \cite{Dong_2019_CVPR}.

Adversarial training constitutes the current state-of-the-art method to defend against adversarial attacks. The key idea is to train a target model on adversarial examples and corresponding class labels at the same time. It can be interpreted as a new kind of data augmentation approaches. Previous works \cite{42503,madry2017towards,tramer2018ensemble} have shown that adversarial training can effectively increase the adversarial robustness of deep neural networks, especially against white-box attacks. However, most existing adversarial training methods could be also problematic. First, there is a risk of overfitting to the perturbations crafted with a specific attack \cite{tramer2018ensemble}, thus the trained model may not generalize well to adversarial examples from other attacks and different test datasets. Second, standard accuracy on clean images often drops as a result of adversarial training, which is undesirable. The trade-off between adversarial robustness and standard accuracy must be taken into account when designing defense methods against adversarial attacks.

In this paper, we propose a novel defense method against adversarial attacks from the perspective of domain adaptation. This is motivated by the observation that there is a considerable distribution mismatch between the clean images and adversarial examples in the high-level feature space. The adversarial perturbations can be progressively amplified to a large magnitude in higher layers of a target model and eventually result in wrong predictions \cite{liao2018defense}. To mitigate the domain shift, we propose a \textbf{C}lass-\textbf{A}ware \textbf{D}omain \textbf{A}daptation (CADA) approach to improve the robustness of deep convolutional networks against various adversarial attacks. In particular, we propose to train a discriminator for unsupervised domain adaptation to minimize the $\mathcal{H}$-divergence \cite{ben2007analysis} between clean images and adversarial examples. In this way, the trained models can learn robust features that are domain-invariant, which can effectively improve the adversarial robustness. Moreover, we incorporate a class-aware component into the discriminator to exploit the label information of adversarial examples. It can increase the discriminative power of the target model for  adversarial examples.

The main contribution of this work can be summarized as follows:

\begin{itemize}
   \item We formulate the adversarial defense as a domain adaptation problem and propose a novel Class-Aware Domain Adaptation framework to improve the adversarial robustness of deep neural networks.

   \item We achieve effective adversarial defense by using a domain discriminator to reduce the distribution mismatch between the representations of adversarial examples and clean images. Furthermore, we make the discriminator class-aware by exploiting the label information of adversarial examples, which helps the target model to learn more discriminative features for adversarial examples.

   \item Evaluated on different benchmark datasets, our method significantly improves the state-of-the-art in terms of adversarial robustness against different attacks and also maintains high accuracy on clean images.

\end{itemize}


\section{Related Work}
\label{sec:related_work}
In this section, we first specify some of the notations used in this paper and then provide a brief review of current methods on adversarial attacks and defenses.

\subsection{Notation}
In this paper, we use $x$ to denote the clean image in a dataset (either train or test set) and $y$ to denote the class label. The ground-truth is denoted by $y_{true}$. We consider neural network based classifier $f: x \rightarrow y$ as our target model. For convenience, let $\phi(x)$ be the feature extraction mapping (the last layer of neural network before fully connected layer) of the input image $x$. $J(f(x), y)$ is used to denote the loss function of classification. $x^{adv}$ denotes the adversarial example generated by perturbing the clean image $x$. $\epsilon$ is the magnitude of the adversarial perturbation.

\subsection{Attacks}
Attacks are commonly divided into two types based on the amount of information that an attacker can obtain: \emph{White-box attacks} have the full information about the target model including architecture, parameters, gradients, \emph{etc}. \emph{Black-box attacks} have very limited knowledge about the target model and cannot send queries to obtain more information.

The seminal work \cite{42503} first shows that deep neural networks are vulnerable to adversarial examples and demonstrates adversarial attacks can be transferred across different deep learning models. Since then, more adversarial attacks have been proposed to fool deep convolutional neural networks. Fast Gradient Sign Method (FGSM) \cite{43405} is introduced to attack a classifier by using the gradients of the loss with respect to the input images. This method tries to maximize the loss function to find the adversarial examples by running gradient ascent for one iteration:
\begin{equation}
x^{adv} = x + \epsilon \cdot \textrm{sign}(\nabla_x J(f(x), y_{true}))
\end{equation}
As a single step attack, FGSM is able to generate adversarial examples efficiently. Following works \cite{kurakin2016adversarial} discover the label leaking effect of FGSM and suggest to replace ground-truth labels with the predicted ones of the target model.

In addition, iterative attacks are introduced to provide much stronger attacks. Basic Iterative Method (BIM) uses a similar formulation as FGSM, however it runs the gradient ascent optimization for multiple iterations and applies small perturbations in each iteration \cite{kurakin2016adversarial}. Using subscript to denote the iteration number and $\alpha$ for the attack step, BIM can be formulated as
\begin{align}
x_0^{adv} &= x \\
x_{t+1}^{adv} &= x_t^{adv} + \alpha \cdot \textrm{sign}(\nabla_{x_t^{adv}} J(f(x_t^{adv}), y_{true}))\\
x_{t+1}^{adv} &= \textbf{\textrm{clip}}(x_{t+1}^{adv}, x_{t+1}^{adv} - \epsilon, x_{t+1}^{adv} + \epsilon)
\end{align}
where $\textbf{\textrm{clip}}(\cdot, a, b)$ makes sure its input to reside in the range $[a, b]$. 

Projected Gradient Descent (PGD) attack \cite{madry2017towards} is a similar iterative attack by projecting the perturbed images into the feasible solution space, which ensures a maximum per-pixel perturbation being no greater than $\epsilon$ (that is subject to an $L_{\infty}$). Different from BIM, PGD consists of initializing search at a random point within a limited area of the original images. As a result, the noisy initialized point can help create a much stronger attack than previous iterative methods. Moreover, PGD has been widely used to investigate the adversarial robustness of deep neural networks.

Recently other types of adversarial attacks are also proposed to fool deep neural networks. One pixel attack \cite{su2019one} is introduced by modifying only one pixel based on differential evolution. Simple rotating 2D images \cite{engstrom2017rotation} are also used to fool neural network-based vision systems. Translation-invariant
attack \cite{dong2019evading} is proposed to produce more transferable adversarial examples, and adversarial examples in the physical world can be produced by printing the digitally perturbed image on paper \cite{kurakin2017adversarial}. Moreover, adversarial attacks have been used to fool the object detector \cite{wang2020adversarial}.

\subsection{Defenses}
With the advancement of attack techniques, many adversarial defense techniques have been also developed. Adversarial training \cite{43405,tramer2018ensemble,madry2017towards} is one of the most investigated defense methods against adversarial attacks by augmenting the training dataset with adversarial examples. Adversarial training can effectively improve the adversarial robustness of the target model by learning the perturbation pattern. 

FGSM adversaries are first used by adversarial training to defend a single step adversarial attack \cite{43405}. Other works \cite{madry2017towards,tramer2018ensemble} try to improve the robustness against iterative attacks such as BIM and PGD. Adversarial Logit Pairing (ALP) is proposed to enhance adversarial training by encouraging the logit predictions of a network for a clean image and its adversarial counterpart to be similar \cite{kannan2018adversarial}. A feature denoising architecture \cite{xie2019feature} is proposed to further improve the effectiveness of adversarial training. A new defense method, TRADES \cite{zhang2019theoretically}, is designed with a theoretical analysis of the trade-off between accuracy and adversarial robustness. It is also possible to achieve adversarial defenses by using Generative Adversarial Networks (GANs) \cite{liu2019gandef}, knowledge distillation \cite{papernot2016distillation}, high-level representation guided denoiser \cite{liao2018defense}, image patches denoising \cite{moosavi2018divide}, noise injection \cite{He_2019_CVPR} and convolutional sparse coding \cite{Sun_2019_CVPR}. More recently, ATDA \cite{song2018improving} is proposed to improve the generalization of adversarial training by minimizing the gap between the clean images and adversarial examples. Their work forces the similarity between the logits predictions of clean images and adversarial examples through conventional methods such as covariance matrix alignment. In contrast, our model focuses on reducing the distribution mismatch between the distributions of adversarial examples and clean images in the feature space by a domain discriminator.

\begin{figure*}[!tb]
\begin{center}
\includegraphics[width=\textwidth]{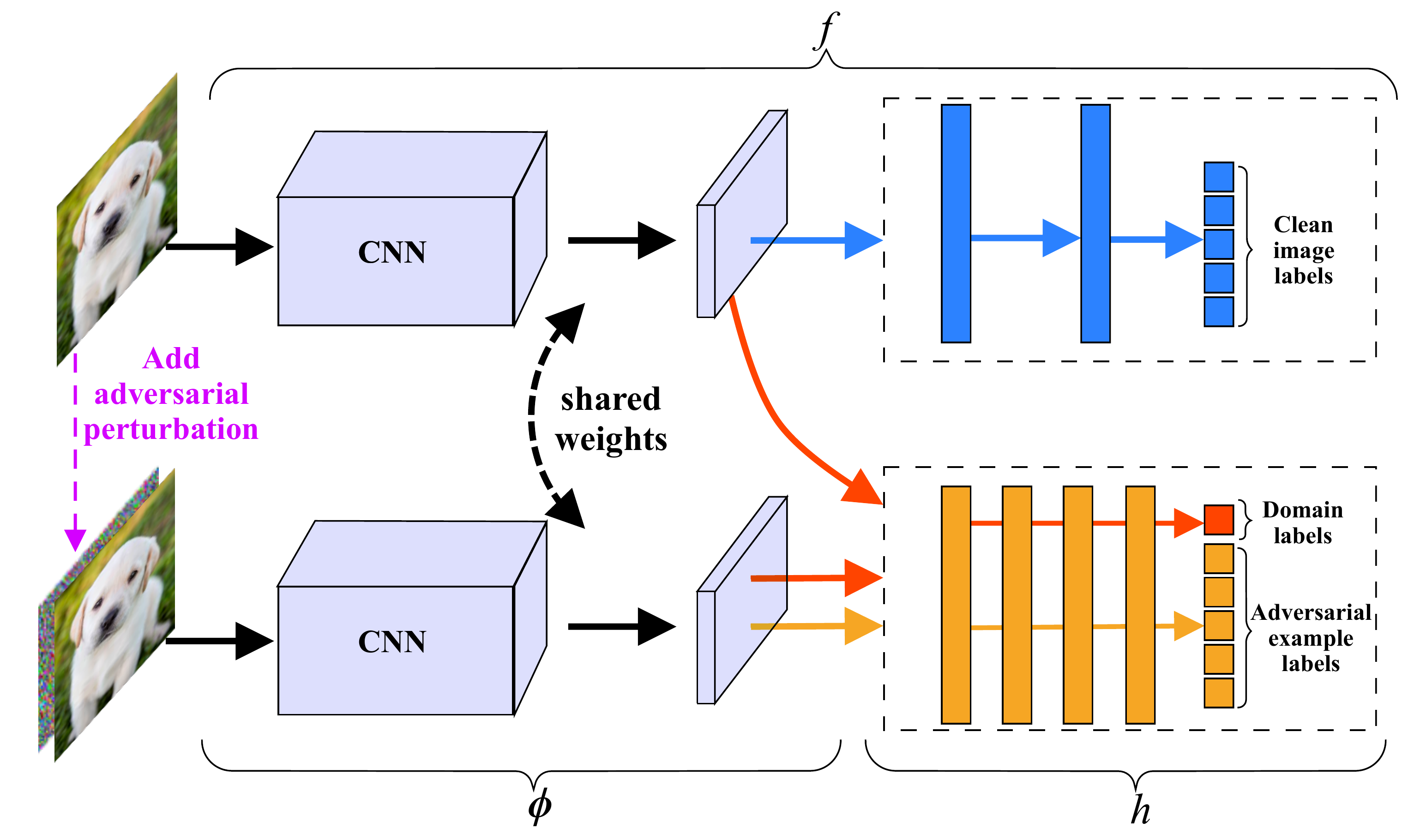}
\end{center} 
   \caption{Overview of our model. $f$ is a conventional image classification network. The left part is a stack of convolutional blocks $\phi$ for feature extraction and is shared for both clean and adversarial images. The bottom right is our discriminator with 4 fully connected layers.
   }
\label{fig:overview}
\end{figure*}

\section{Our Method}
\label{sec:method}
We propose a Class-Aware Domain Adaptation method to deal with adversarial attacks from the perspective of domain adaptation. Specifically, we use a domain discriminator to align the feature distributions of clean images and adversarial examples based on $\mathcal{H}$-divergence \cite{ben2010theory} theory and implement it based on generative adversarial training \cite{goodfellow2014generative}. Moreover, we make the discriminator class-aware to increase the discriminative power of the target model for adversarial images.

\subsection{Distribution Alignment}
$\mathcal{H}$-divergence \cite{ben2007analysis,ben2010theory,chen2018domain} is used to measure the divergence between two sets of data with different distributions. Let $z_s$ and $z_t$ be the distributions from source domain ($\mathcal{D}$) and target domain ($\mathcal{D'}$) respectively. We denote a labeling function $h: z \rightarrow [0, 1] $ as a domain classifier and $\mathcal{H}$ as a set of possible domain classifiers, where $h \in \mathcal{H}$. Therefore, the $\mathcal{H}$-divergence between $D$ and $D'$ is:
\begin{equation}
d_{\mathcal{H}}(\mathcal{D}, \mathcal{D'}) =2 \Big(1 - \min_{h \in \mathcal{H}} \Big[ \frac{1}{N} \sum_z \mathcal{L}_{\mathcal{D}} (h(z))+ \frac{1}{N} \sum_{z}\mathcal{L}_{\mathcal{D'}}(h(z))\Big]\Big)
\end{equation}
where $\mathcal{L}_{\mathcal{D}}$ and $\mathcal{L}_{\mathcal{D'}}$ denote the prediction errors (cross-entropy loss) of $h(z)$ on source and target domain respectively. $N$ is the number of samples for a given dataset.

Under the context of deep learning, $z$ represents the feature representation of an input image $x$ extracted from a neural network, \emph{i.e.}, $z = \phi(x)$. In order to align the distributions between $D$ and $D'$, we need to optimize $\phi$ to output feature representations that minimize the domain divergence $d_{\mathcal{H}}(\mathcal{D}, \mathcal{D'})$, which leads to a minimax game between $\phi$ and $h$:
\begin{equation} \min_{\phi}{d_{\mathcal{H}}(\mathcal{D}, \mathcal{D'})} \Leftrightarrow \max_{\phi} \min_{h \in \mathcal{H}} \Big[ \frac{1}{N} \sum_z \mathcal{L}_{\mathcal{D}} (h(z))+ \frac{1}{N} \sum_{z}\mathcal{L}_{\mathcal{D'}}(h(z))\Big]
\end{equation}

\subsection{Domain Adaptation}
We seek to improve the adversarial robustness of deep neural networks from the perspective of domain adaptation. It is achieved by reducing the distribution mismatch between clean images and adversarial examples on the feature space based on $\mathcal{H}$-divergence theory above. In particular, we refer to the domain of clean images as source domain and the domain of adversarial examples as target domain. Our discriminator produces probability distributions over domain labels, $h_d : z \rightarrow h_{d}(z)$. In addition, we follow LSGAN \cite{mao2017least} to replace the negative log likelihood objective by a least square loss. Thus, the domain adaptation objective can be formulated as:
\begin{equation}
\mathcal{L}_{ada}(\phi, h_{d}) = \frac{1}{N} \Big[\sum_x \big(h_{d}(\phi(x)) - 1\big)^2 + \sum_{x^{adv}} h_{d}(\phi(x^{adv}))^2\Big]
\end{equation}
where $\phi$ tries to output similar feature representations $\phi(x^{adv})$ and $\phi(x)$ for both adversarial examples and clean images, while $h_d$ aims to distinguish between $\phi(x^{adv})$ and $\phi(x)$. In other words, $\phi$ tries to minimize this objective, while $h_d$ tries to maximize it.

\subsection{Class-Aware Discriminator}
Even though the binary discriminator has the capability of aligning all the clean images and adversarial examples into a similar feature space, it could suffer from the problem of mode collapse and there is no guarantee that the adversarial examples with the same class label would be mapped nearby in the feature space. To this end, we add an auxiliary classifier on top of the binary discriminator by utilizing the label information in the adversarial domain. As a result, it can help the adversarial examples to preserve their categorical information. Our class-aware discriminator can produce the probability over both domain labels as well as class labels for adversarial examples,  $h : z \rightarrow \{h_{d}(z), h_{c}(z)\}$. Thus, the classification loss of adversarial examples is defined as:
\begin{equation}
\mathcal{L}_{cls}^{adv}(\phi, h_{c}) = \frac{1}{N} \sum_{x^{adv}} J\big(h_{c}(\phi(x^{adv})), y_{true}\big)
\end{equation}

\subsection{Full Objective}
By defining the classification loss of the clean images as:
\begin{equation}
\mathcal{L}_{cls}(f) = \frac{1}{N} \sum_{x} J\big(f(x), y_{true}\big)
\end{equation}
our full objective is:
\begin{equation}
\label{eq:objective}
\mathcal{L}(f, \phi, h_{c}, h_{d}) = \mathcal{L}_{cls}(f) + \lambda_{1} \mathcal{L}_{cls}^{adv}(\phi, h_{c}) + \lambda_{2} \mathcal{L}_{ada}(\phi, h_{d})
\end{equation}
where the $\lambda_{1}$ and $\lambda_{2}$ control the relative importance of the different objectives. We aim to solve:
\begin{equation}
\label{equ:minimax}
f^*, h^* = \textrm{arg} \min_{f,\phi,h_c} \max_{h_d} \mathcal{L}(f, \phi, h_c, h_d)
\end{equation}
Notice that $\phi$ is part of $f$ for feature extraction. Discriminator $h$ consists of $h_d$ and $h_c$, which share all the parameters except the last fully connected layer. We achieve the joint optimization of Equation~\ref{equ:minimax} by switching domain labels between clean images and adversarial examples similar to GAN training \cite{goodfellow2014generative}.

\subsection{Network Overview}
The overview of our network is shown in Figure~\ref{fig:overview}. Our class-aware discriminator can be integrated with a conventional image classification network $f$, which is illustrated in the upper part of Figure \ref{fig:overview}. The left part is a stack of convolutional blocks $\phi$ for feature extraction and is shared for both clean and adversarial images. The bottom right is our discriminator which consists of 4 fully connected layers and is added after the feature extraction blocks. The dimension of all the hidden layers of the discriminator is set as the same as that of the extracted features. The whole system can be trained in an end-to-end manner using SGD optimization. During testing phase, the discriminator can be removed and we can use the target classification network with robust weights.

\section{Experiment}
\label{sec:experiment}
In this section, we evaluate CADA on various widely used benchmarks and compare against recent methods for adversarial defenses. Experimental analysis and ablation study are also provided.

\subsection{Experiment Setup}
Following previous works \cite{madry2017towards,kannan2018adversarial}, we use PGD to generate adversarial examples on-the-fly in every training iteration. We evaluate the classification accuracy on test images that are adversarially perturbed by FGSM, BIM and PGD respectively. In this work, all the attacks are considered to consist of perturbations of limited $L_{\infty}$ norm with an allowed maximum value of $\epsilon$. Adversarial Robustness Toolbox \cite{art2018} is used to produce adversarial examples.

We evaluate the proposed approach on MNIST, CIFAR10 and CIFAR100 datasets under both the white-box and black-box attack settings.  For black-box attacks, we evaluate the trained model with adversarial examples transferred from a copy of the same classification model, which is independently initialized and trained.

For all the experiments, the image pixel values are normalized to [0, 1]. We set the weighting parameters $\lambda_1 = 0.5$, $\lambda_2 = 0.5$ to train MNIST and CIFAR10, and use $\lambda_1 = 0.5$, $\lambda_2 = 1.0$ for CIFAR100. All the models are trained by using Adam \cite{kingma2015adam} for stochastic optimization with a batch size of 64 and a total of 250 epochs. The initial learning rate is set to 3$\times 10^{-4}$ and decreased by 10$\times$ at the $150^{th}$ epoch. Our implementation is based on deep learning framework PyTorch with a single GTX 1080Ti GPU.

\begin{table*}[!tb]
\centering
\tabcolsep=0.1cm
\caption{Test accuracy on clean images and adversarial examples under various white-box and black-box attacks. We set the total adversarial perturbation $\epsilon=0.3$ and number of iterations $K=40$ for BIM and PGD attacks on MNIST datasets. 
We set $\epsilon=0.031$ and $K=20$ for CIFAR10 and CIFAR100 datasets. The perturbation of FGSM is the same as BIM and PGD. }
\begin{tabular}{cc|c|ccc|ccc}

\\
\hline
\multirow{2}{*}{Dataset}  & \multirow{2}{*}{Defense} & \multirow{2}{*}{Clean \newline(\%)} & \multicolumn{3}{c|}{White-box attack (\%)} & \multicolumn{3}{c}{Black-box attack (\%)} \\ \cline{4-9} 
                          &                          &                             & FGSM         & BIM          & PGD         & FGSM         & BIM          & PGD         \\ \hline
\multirow{6}{*}{MNIST}    & Vanilla                  & \textbf{99.16}                       & 14.06        & 0.74         & 0.75        & 14.10        & 0.74         & 0.74        \\  
                          & AT                       & 99.15                       & 96.63        & 71.87        & 94.56       & 97.13        & 78.52        & 95.69       \\  
                          & ALP                      & 98.30                       & 96.40        & 88.26        & 96.11       & 96.64        & 89.91        & 96.27       \\  
                          & TRADES                   & 98.95                       & 96.88        & 89.00        & 95.79       & 96.91        & 91.64        & 96.01       \\  
                          & CADA                     & 98.80                       & \textbf{98.57}        & \textbf{98.52}        & \textbf{98.09}       & \textbf{98.62}        & \textbf{98.84}        & \textbf{98.56}       \\ \hline
\multirow{6}{*}{CIFAR10}  & Vanilla                  & \textbf{92.66}                       & 25.65        & 9.26         & 7.31        & 44.34        & 15.15        & 18.42       \\  
                          & AT                       & 86.28                       & 63.05        & 33.18        & 48.96       & 75.01        & 57.32        & 72.68       \\  
                          & ALP                      & 83.82                       & 62.07        & 33.51        & 53.21       & 72.47        & 50.50        & 71.88       \\  
                          & TRADES                   & 84.22                       & 69.54        & 40.50        & 54.85       & 75.26        & 64.58        & 73.05       \\  
                          & CADA                     & 86.55                       & \textbf{73.93}        & \textbf{71.31}        & \textbf{71.59}       & \textbf{78.00}        & \textbf{75.43}        & \textbf{78.37}       \\ \hline
\multirow{6}{*}{CIFAR100} & Vanilla                  & \textbf{72.46}                       & 25.68        & 11.15        & 9.16        & 37.01        & 35.88        & 33.48       \\  
                          & AT                       & 57.58                       & 33.53        & 19.82        & 27.91       & 48.05        & 37.72        & 48.25       \\  
                          & ALP                      & 58.20                       & 35.52        & 22.02        & 28.96       & 46.20        & 37.60        & 46.61       \\  
                          & TRADES                   & 59.32                       & 36.51        & 25.92        & 29.29       & 48.61        & 36.88        & 48.46       \\  
\                          & CADA                     & 67.81                       & \textbf{41.69}        & \textbf{26.73}        & \textbf{29.42}       & \textbf{49.84}        & \textbf{40.53}        & \textbf{48.63}       \\ \hline
\end{tabular}
\label{tab:results}

\end{table*}

\begin{table}[!tb]
\centering
\tabcolsep=0.1cm
\caption{The comparisons of CADA with ATDA \cite{song2018improving} on CIFAR10 and CIFAR100 test dataset. The total adversarial perturbation $\epsilon=4/255$ and number of iterations $K=20$.\newline}
\scalebox{0.8}{
\begin{tabular}{cc|c|cc|cc}
\hline
\multirow{2}{*}{Dataset}  & \multirow{2}{*}{Defense} & \multirow{2}{*}{Clean (\%)} & \multicolumn{2}{c|}{White-box attack (\%)} & \multicolumn{2}{c}{Black-box attack (\%)} \\ \cline{4-7} 
                          &                          &                            & FGSM                 & PGD                 & FGSM                 & PGD                 \\ \hline
\multirow{2}{*}{CIFAR10}  & ATDA                     & 84.8                       & 60.7                 & 58.1                & 80.7                 & 80.7                \\ 
                          & CADA                     & 86.5                       & 76.3                 & 73.8                & 82.7                 & 82.6                \\ \hline
\multirow{2}{*}{CIFAR100} & ATDA                     & 61.6                       & 29.3                 & 26.2                & 56.0                 & 56.0                \\ 
                          & CADA                     & 63.9                       & 36.1                 & 30.3                & 57.7                & 57.9                \\ \hline
\end{tabular}
}
\label{tab:atda}

\end{table}

\subsection{Experimental Results}
We compare our CADA approach with state-of-the-art adversarial defense methods ALP \cite{kannan2018adversarial} and TRADES \cite{zhang2019theoretically}. In addition, we use Vanilla to represent the model trained with clean images, and AT to denote the Adversarial Training \cite{madry2017towards} with both the clean images and adversarially perturbed images produced by PGD attack.

\subsubsection{Results on MNIST.} The LeNet \cite{lecun1998gradient} is used as the target classification model. We set the total adversarial perturbation $\epsilon = 0.3$, perturbation step size $\eta = 0.01$ and number of iterations $K = 40$ for training. To evaluate the robust accuracy on adversarial images, we apply BIM and PGD attacks with $K = 40$ iterations and $\eta = 0.01$ step size.  The results are summarized in the upper part of Table~\ref{tab:results}. We can see that although all the defense methods are able to retain high accuracy on clean images, our CADA approach generalizes much better on adversarial examples. In particular, we can achieve over 98\% robust accuracy for all the attacks.

\subsubsection{Results on CIFAR10.} We apply the pre-activation ResNet-18 \cite{he2016identity} for image classification. We set the total perturbation $\epsilon = 0.031$, perturbation step size $\eta = 0.003$ and number of iterations $K = 10$ for training. For evaluation, we apply BIM and PGD attacks with $K = 20$ iterations and $\epsilon = 0.031$ perturbations. The results of the test accuracy are shown in the middle part of Table~\ref{tab:results}. It shows that our CADA method is significantly better than other defense methods. We can achieve more than 70\% accuracy for all the white-box attacks, while ALP can only achieve 62.07\%, 33.51\% and 53.21\% accuracy on FGSM, BIM and PGD adversaries respectively. It also shows that CADA can outperform all the competing methods on all the adversaries under the black-box setting. Furthermore, we evaluate our method under the same attack setting as ATDA \cite{song2018improving}. As shown in Table~\ref{tab:atda}, our method can outperform ATDA by a clear margin.

\subsubsection{Results on CIFAR100.} We also use the pre-activation ResNet-18 as classification model for CIFAR100 dataset and the output size of the last layer is changed to 100. We adopt the training settings as the same as CIFAR10 dataset, and apply BIM and PGD attacks with $K = 20$ iterations and $\epsilon = 0.031$ perturbations for evaluation. The results are shown in the bottom part of Table~\ref{tab:results}. Moreover, the comparison with ATDA is shown in Table~\ref{tab:atda}. It is clear that our CADA method outperforms other methods under both white-box and black-box settings.

\subsubsection{Adversarial Robustness v.s. Standard Accuracy.} We further investigate the trade-off between the robustness on adversarial examples and standard accuracy on clean images. As shown in the third column of Table~\ref{tab:results}, training robust models with adversarial examples generally lead to a reduction of standard accuracy on clean images. However, it can be seen that our models can retain relatively higher accuracy on clean images than other defense methods. In particular, our model only has a slight reduction in clean accuracy to 67.81\% (72.46\% for Vanilla training) on CIFAR100 images, while there are more than 13\% accuracy drop for other methods. The reason is that our robust models are not directly trained on classification loss of adversarial examples (see Equation \ref{eq:objective}). The labels of adversarial examples are only served as a kind of regularization via the discriminator, thus reducing the risk of overfitting on adversarial perturbations for the target model.

\begin{figure*}
\begin{center}
\includegraphics[width=\textwidth]{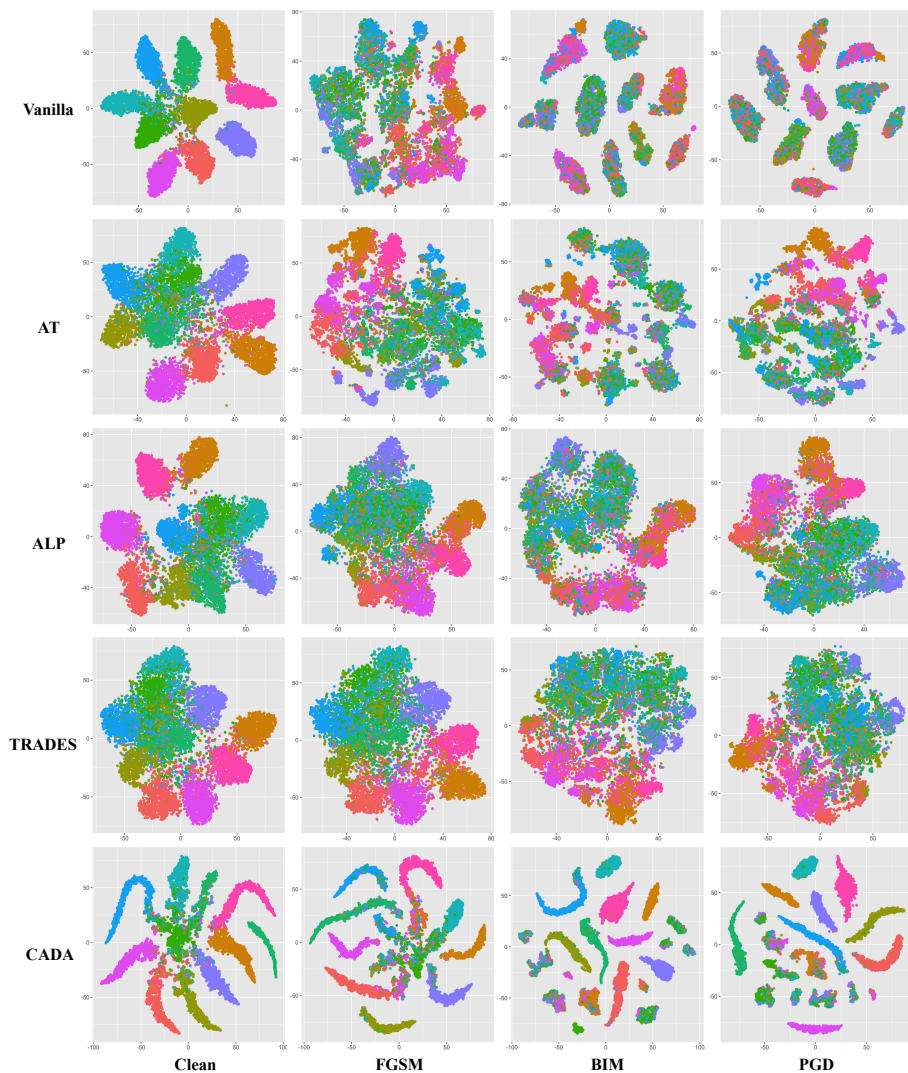}
\end{center} 
   \caption{Feature visualizations by t-SNE for CIFAR10 test images. Vanilla in the first row represents the results with standard training. The second to the fifth row correspond to different defense methods AT, ALP, TRADES and CADA respectively. The first to the fourth column correspond to clean images as well as adversarial examples produced by the white-box FGSM, BIM and PGD attacks.}
\label{fig:tsne}
\end{figure*}

\subsubsection{Different Types of Attacks.} Although all the defense methods are only trained with PGD adversaries, our method generalizes much better across different types of adversarial attacks. As shown in Table~\ref{tab:results}, we can see that ALP and TRADES are able to achieve 96.11\% and 95.79\% accuracy on MNIST dataset with PGD adversaries under white-box setting, however they can only reach 88.26\% and 89.00\% accuracy with BIM attack respectively. In contrast, our CADA can achieve robust accuracy as high as 98.52\% under the same setting, and generalizes well for black-box attacks. Moreover, the superiority of our method can be also observed when evaluating CIFAR images. In particular, our method can achieve similar accuracy around 71\% under both BIM and PGD white-box attacks for CIFAR10 images. However the accuracy drops considerably for other defense methods when tested on BIM adversaries, \emph{e.g.,} decreasing from 53.21\% to 33.51\% for ALP.




\subsection{Distribution Visualization}
We further investigate the distribution of the feature representations produced by different defense methods. In particular, we use t-Distributed Stochastic Neighbor Embedding (t-SNE) \cite{maaten2008visualizing} to visualize the structure of the high-dimensional image representations by giving each image a location in a two-dimensional map. t-SNE is capable of arranging images that have similar high-dimensional vectors to be nearby in the embedded space. We consider different defense methods as well as adversarial examples produced by various white-box attacks.

As shown in Figure~\ref{fig:tsne}, the extracted features of CIFAR10 test images are visualized as a scatterplot in which the colors represent the labels of different categories. The first row shows the results of Vanilla training with only clean images. We can see that the clean images can be well separated into their natural clusters and the adversarial examples are clustered with the wrong classes.

\begin{figure}[!tb]
\begin{center}
\includegraphics[width=\textwidth]{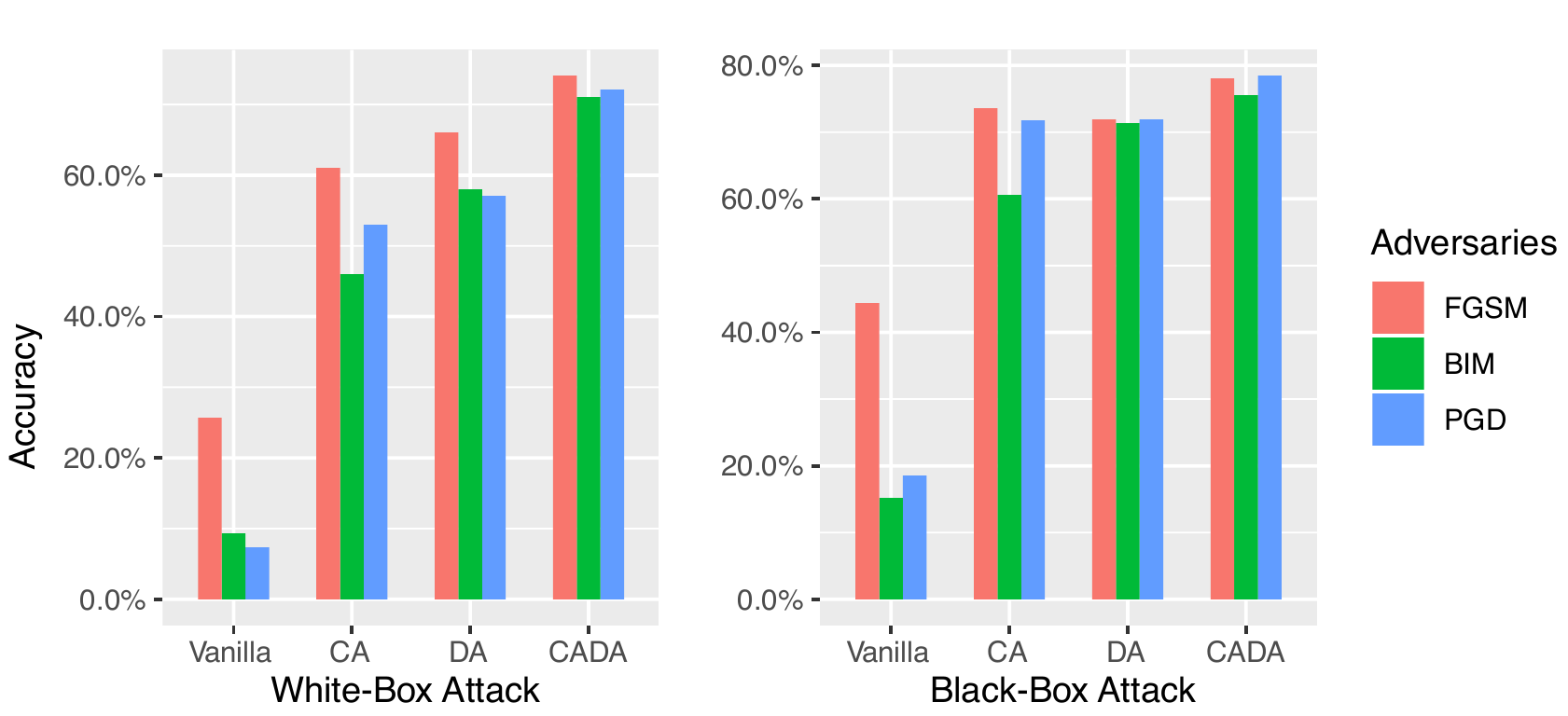}
\end{center} 
   \caption{Performance comparison of different variants of our model on CIFAR10. We compare the classification accuracies on the test set with FGSM, BIM and PGD attacks under both white-box and black-box settings.
   }
\label{fig:cifar10_ablation}
\end{figure}
\begin{figure}[!tb]
\begin{center}
\includegraphics[width=\textwidth]{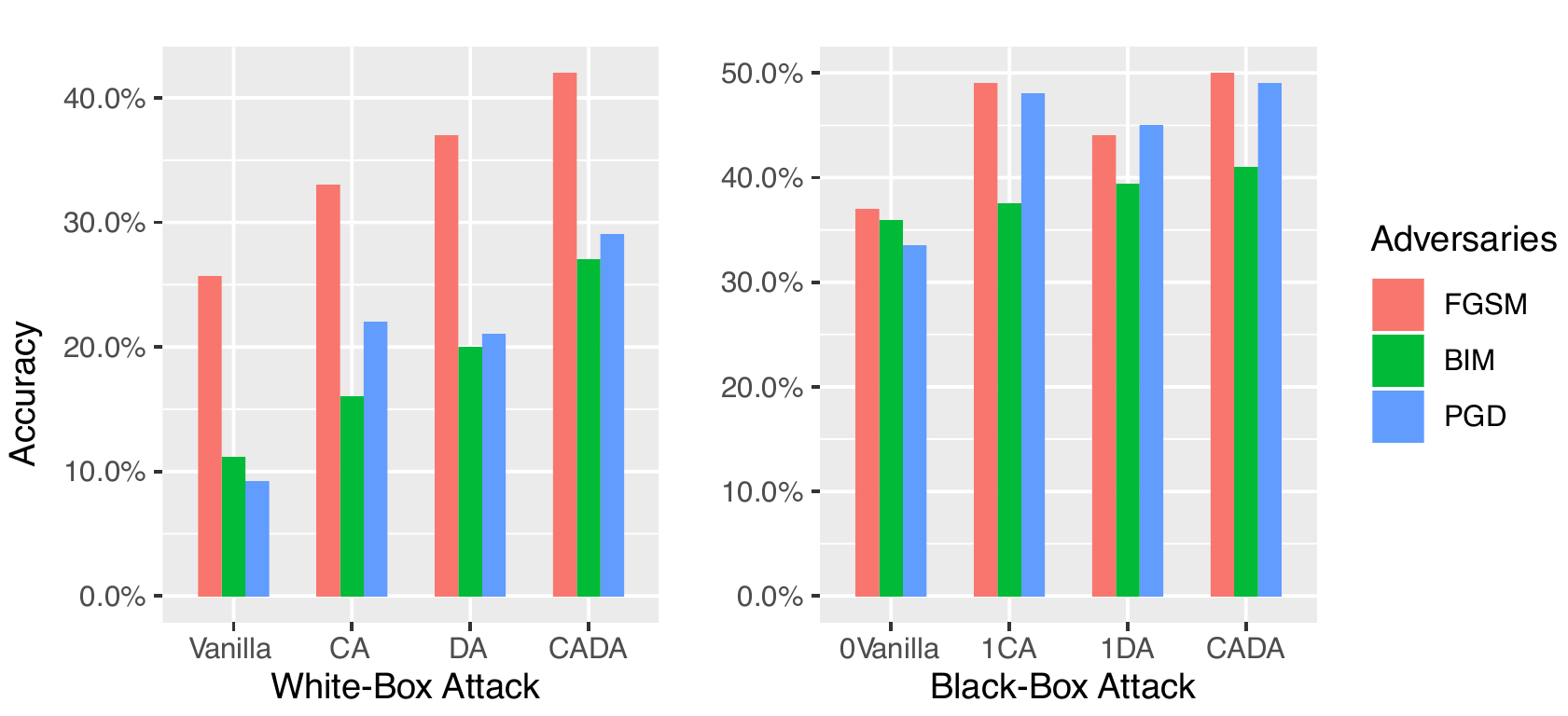}
\end{center} 
   \caption{Performance comparison of different variants of our model on CIFAR100. We compare the classification accuracies on the test set with FGSM, BIM and PGD attacks under both white-box and black-box settings.
   }
\label{fig:cifar100_ablation}
\end{figure}

The second to the fourth row correspond to the results of 3 different defense methods AT, ALP and TRADES. It can be seen that the maps constructed for clean images are significantly better and most of the members of each class fairly close together. However the results produced for adversarial examples provide little insight into the class structure of the dataset since there are large overlaps between different classes. In contrast, the t-SNE visualization based on our CADA is able to construct a map in which the separation between different classes is almost perfect. It is clear that our method does a much better job to reveal the categorical information of the CIFAR10 dataset. Moreover, our t-SNE maps exhibit the similar distribution structures for both clean images and different adversarial examples, demonstrating that our method can learn domain-invariant features for both clean images and adversarial examples. As a result, our method is able to balance well between adversarial robustness and standard accuracy.


\subsection{Ablation Study}
We demonstrate the necessity of the two components of our method by comparing the adversarial robustness of several ablated versions of CADA. We use DA to denote the trained model with Domain Adaptation component only and CA to represent the model with Class-Aware component only.

The ablation results are shown in Figure~\ref{fig:cifar10_ablation} and Figure~\ref{fig:cifar100_ablation} for CIFAR10 and CIFAR100 respectively. Specifically, compared with Vanilla training, the classification performance can be significantly improved with both CA and DA. The improvements generalize well across different adversarial attacks under both white-box and black-box settings. This proves that both domain adaptation component and class-aware component can effectively improve the adversarial robustness of classification models, and the best performances are achieved with CADA by combining the two components. Moreover, the high performance of DA demonstrates that the adversarial robustness can be significantly improved even without the labels of the adversarial examples.

We also observe that there are several obvious performance drops for CA under BIM attack for both CIFAR10 and CIFAR100 images, while DA exhibits more stable improvements under both BIM and PGD attacks. Considering that all the models are trained with PGD adversaries, it can be concluded that domain adaptation component plays a key role in learning domain-invariant features and generalizing well on various attacks.

\begin{figure}[!tb]
\includegraphics[width=\textwidth]{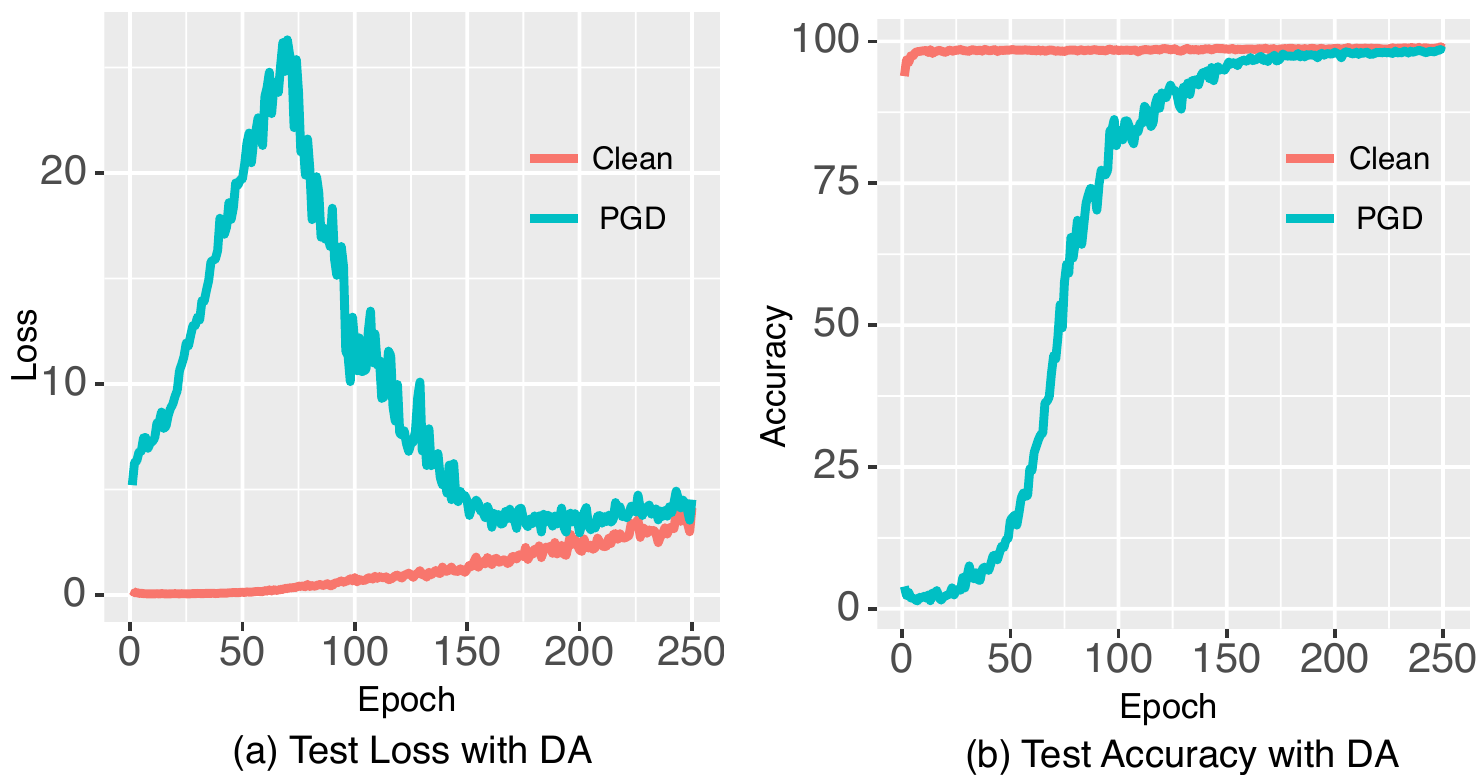}
   \caption{Test loss and accuracy for clean images and PGD adversaries with domain adaptation component on MNIST dataset.
   }
\label{fig:mnist_gan}
\end{figure}

Furthermore, we provide an empirical study on the evolution of the effectiveness of DA model over time. We calculate the loss and accuracy for clean images and PGD adversaries on MNIST test set and plot them against epochs in Figure~\ref{fig:mnist_gan}. We can find that the loss of PGD adversaries increases rapidly in the early stage of training, then it decreases over time. The robust accuracy of adversaries consistently increases and finally converges to a high value. In contrast, the loss of clean images keeps small but gradually increases a little bit over time, while the standard accuracy converges to a high value after just a few epochs. It is clear that the domain adaptation component can gradually learn domain-invariant features for adversarial examples and clean images. As a result, it helps reduce the risk of overfitting to clean images and improve the adversarial robustness of deep networks.

\begin{figure}[!tb]
\begin{center}
\includegraphics[width=\textwidth]{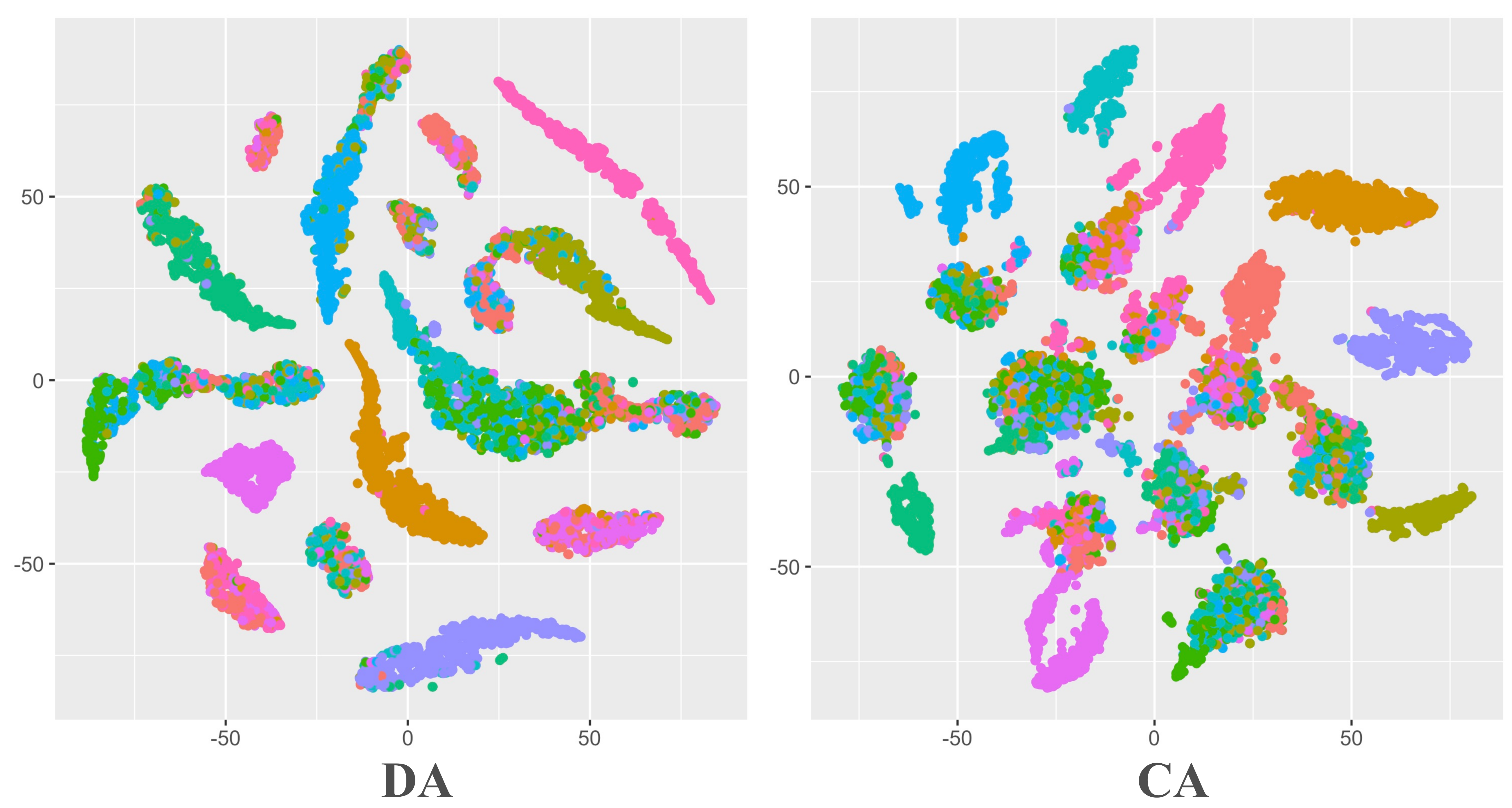}
\end{center} 
   \caption{Feature visualization comparison by t-SNE between Domain Adaptation (DA) and Class-Aware (CA) model. The results are produced on CIFAR10 test images with PGD white-box attack.
   }
\label{fig:cada}
\end{figure}

Finally, we compare the structures of image representations of DA and CA by applying t-SNE to CIFAR10 test dataset. We adopt PGD white-box attack to generate adversarial examples. As shown in Figure~\ref{fig:cada}, the t-SNE maps exhibit different structures for DA and CA. In particular, DA produces a separate map by constructing a ``curve'' shape for the data points of a given class, while CA tends to cluster the images of the same class to nearby points.

\section{Conclusion}
\label{sec:conclustion}
In this paper we have developed a new defense method for improving the adversarial robustness of deep convolutional networks with class-aware domain adaptation. The proposed approach outperforms the state-of-the-art by a large margin in terms of both adversarial robustness and standard accuracy. The experimental results on several benchmark datasets demonstrate that our method also generalizes well across various adversarial attacks under both white-box and black-box settings.

\section*{References}

\bibliography{egbib}

\end{document}